\documentclass[letterpaper, 10 pt, conference]{ieeeconf}  

\IEEEoverridecommandlockouts                              

\usepackage{cite}
\usepackage{amsmath,amssymb,amsfonts}
\usepackage{svg}
\usepackage{graphicx}

\usepackage{subcaption}
\usepackage{textcomp}
\usepackage{xcolor}

\usepackage{hyperref}
\let\labelindent\relax
\usepackage{makecell}
\usepackage{enumitem}
\usepackage{multirow}

\usepackage{algorithm}
\usepackage{algpseudocode}
\usepackage{bbold}
\usepackage[OT1]{fontenc}

\usepackage{hhline}

\title{\LARGE \bf
Grasp-\textcolor{blue}{TAGS}: Task-Aware Grasp Selection via Critic Distillation for
Prehensile Loco-Manipulation}

\author{Xun Tu, Minghao Zou, and Karthik Desingh
\thanks{All authors are affiliated with the Department of Computer Science \& Engineering at the University of Minnesota, Twin Cities.
Contact:{\tt\small \{tu000080, zou00189, kdesingh\}@umn.edu}}%
}

\let\oldtwocolumn\twocolumn
\renewcommand\twocolumn[1][]{%
    \oldtwocolumn[{#1}{
           \centering
           \includegraphics[width=1.0\textwidth]{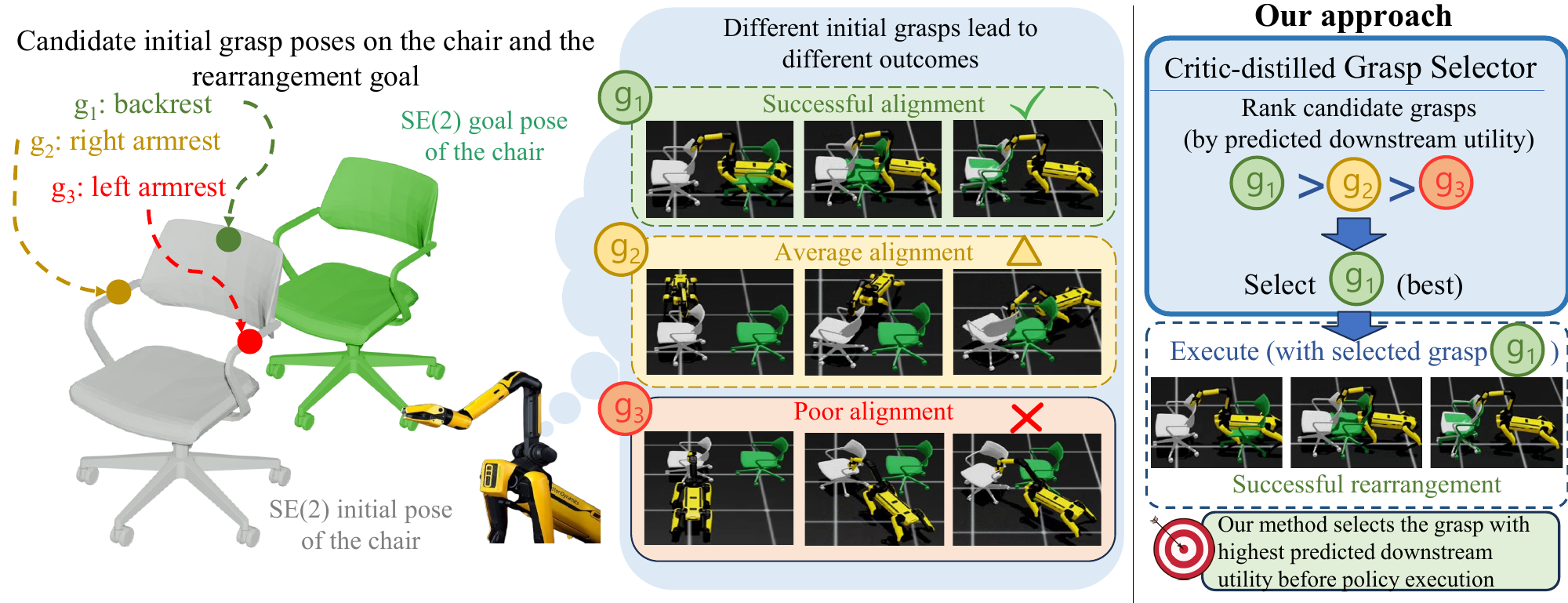}
           \captionof{figure}{\footnotesize{\textbf{Overview of the Grasp-TAGS framework.} Different feasible grasp poses on the same object can induce substantially different downstream loco-manipulation behaviors and rearrangement outcomes. For example, three grasp poses $g_1$, $g_2$, and $g_3$ on the chair shown here can have different behaviors for moving the chair from its initial pose to the final pose. At inference time, our critic-distilled grasp selector ranks candidate grasps by their predicted downstream task utility, selects the highest-utility grasp for the desired object pose, and executes the manipulation policy from the selected grasp configuration.}}
           \label{fig:overview}
    }]
}

\usepackage{etoolbox}
\makeatletter
\patchcmd{\@makecaption}
  {\scshape}
  {}
  {}
  {}
\makeatother

\begin{document}

\maketitle

\begin{abstract}
Loco-manipulation, which requires mobile manipulators to coordinate locomotion and manipulation over long horizons, is a key capability for large-object rearrangement in applications such as service robotics and warehouse automation. Despite recent progress, reliably executing these tasks remains challenging because performance depends on complex robot-object and object-environment interaction dynamics. Existing approaches address this challenge either through model-based methods, which often require accurate dynamics models, or through learning-based methods, which reduce this requirement but typically assume a fixed or predefined initial grasp. However, different geometrically stable grasps can lead to substantially different downstream manipulation performance. Therefore, we propose Grasp-TAGS, a task-aware grasp selection framework for goal-conditioned large-object rearrangement. We train a unified PPO-based reinforcement learning policy across multiple initial grasp configurations and distill the learned critic into a lightweight grasp utility network that ranks candidate grasps using only the candidate grasp configuration and goal information. At inference time, we select the grasp with the highest predicted downstream utility for the downstream policy to complete the task, before even making a contact. Experiments demonstrate that Grasp-TAGS improves rearrangement success over both fixed single-grasp policies and random grasp selection, achieving an average success rate of 89.84\% compared with 79.87\% for the best-performing single-grasp policies, an improvement of 9.97 percentage points. Grasp-TAGS also reduces final orientation errors across all evaluated objects. 
For more qualitative results, refer to the supplementary webpage: {\small \url{https://rpm-lab-umn.github.io/Grasp-TAGS-web-page/}}.

\end{abstract}

\section{Introduction}

Loco-manipulation enables a mobile robot to coordinate base locomotion and manipulation to physically interact with objects over extended horizons and accomplish manipulation tasks. Large-object rearrangement is a particularly challenging instance of loco-manipulation, where bulky or heavy objects such as chairs, carts, and luggage must be transported to desired poses \cite{SuperQGRASP, fu2022deepwbc, mendonca2024realworldrl,li2025robotmoverlearninglargeobjects}. Because such objects are often impractical to lift, the robot must manipulate them while they remain in contact with the ground, resulting in complex physical interactions not only between the robot and object, but also between the object and its surroundings. Managing these coupled interactions throughout the rearrangement process requires sustained contact, coordinated base-arm motion, and robust control throughout prolonged robot-object-environment interaction. Addressing these challenges is therefore critical to enabling reliable loco-manipulation in real-world applications such as warehouse automation, service robotics, and assistive robotics.

Prior work has addressed loco-manipulation skills for large-object rearrangement using model-based approaches such as model predictive control (MPC), which explicitly model system dynamics and generate whole-body motions through analytical planning and control \cite{Li2022MulticontactMPC, Wang2024AnInteractiveNavigationMethod}. These approaches generally rely on sufficiently accurate knowledge of the underlying dynamics. More recently, learning-based methods using reinforcement learning and imitation learning have demonstrated promising performance \cite{alore2026, li2025robotmoverlearninglargeobjects, ravan2024diffusionmobility, ReLIC}. By learning control policies from simulated interactions or demonstrations, these methods reduce the need to explicitly model complex robot-object-environment dynamics during execution. However, existing learning-based approaches typically assume a fixed or predefined initial grasp configuration before downstream manipulation.

\begin{figure}[htbp!]
     \centering
     \includegraphics[width=\columnwidth]{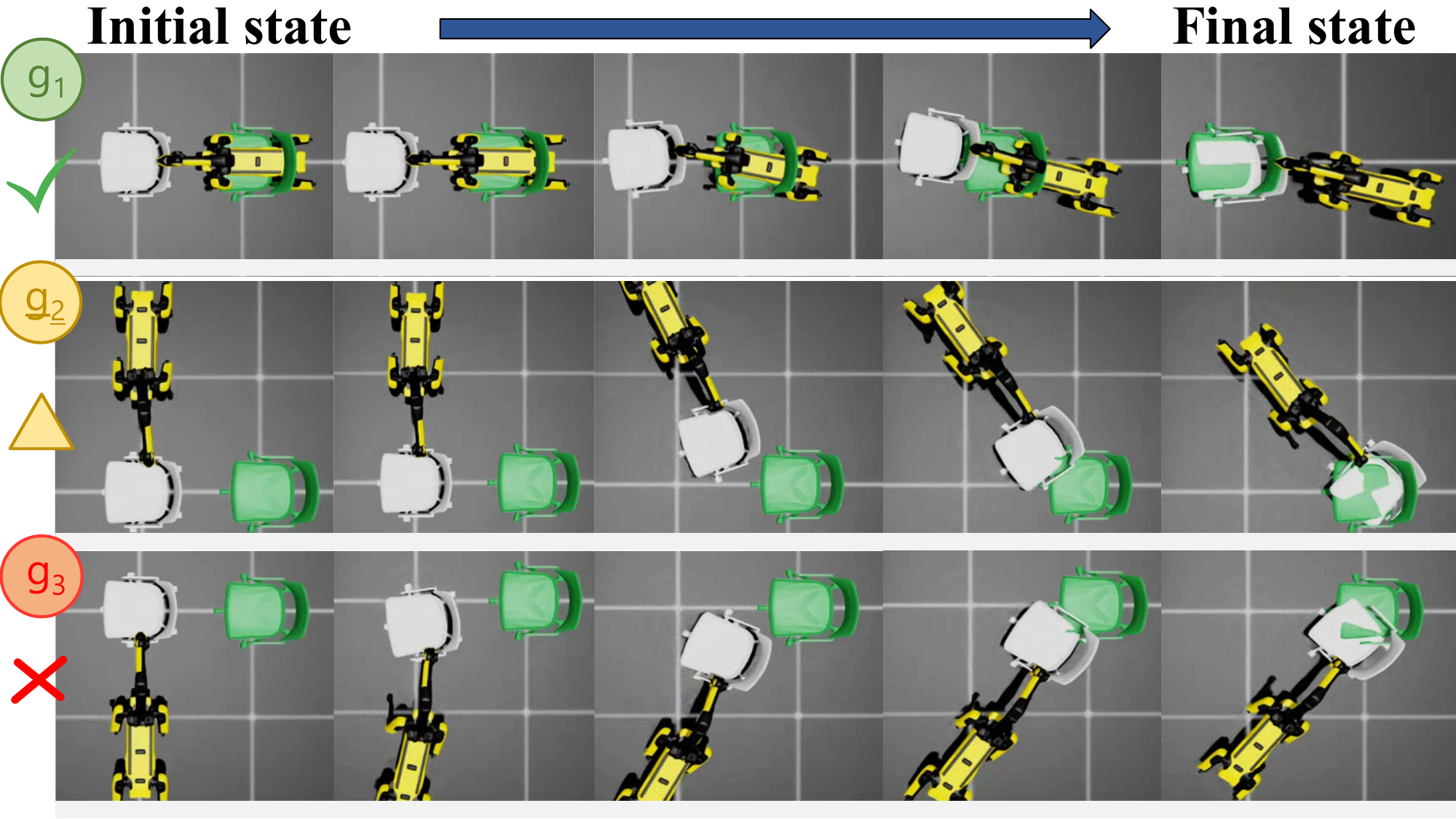}
     \caption{\footnotesize{\textbf{Qualitative example of grasp-dependent rearrangement behavior.} For the illustrated target pose, the backrest grasp ($g_1$) yields better final alignment with the target pose than the two armrest grasps ($g_2$ and $g_3$).}}
     \label{fig:qualitative}
\end{figure}

This assumption is restrictive, as the initial grasp choice can substantially influence subsequent rearrangement. The initial grasp affects downstream rearrangement by changing contact locations and the corresponding arm configuration. These differences alter the motions and forces the robot can effectively apply during pushing or dragging, which in turn influence object controllability and task outcome, as illustrated in Fig.~\ref{fig:qualitative}. 

Consequently, even among geometrically feasible and physically stable grasps, downstream performance can vary substantially with the rearrangement goal. We therefore hypothesize that, for large-object rearrangement, the initial grasp should be treated as a task-dependent decision variable rather than a fixed or random configuration. Specifically, selecting the grasp with the highest expected downstream task utility from multiple feasible candidates should improve rearrangement performance compared with fixed or random grasp selection. 

To this end, we propose Grasp-TAGS, a Task-Aware Grasp Selection framework for goal-conditioned large-object rearrangement, as illustrated in Fig.~\ref{fig:overview}. Grasp-TAGS evaluates candidate grasps according to their expected downstream task utility and selects the most suitable grasp before manipulation begins. We first train a unified manipulation policy via PPO across multiple feasible initial grasp configurations, allowing both the policy and its critic to capture grasp-dependent downstream behavior. Consequently, the critic assigns different value estimates to initial states associated with different candidate grasps. We then distill these grasp-dependent values into a lightweight grasp utility network using only information available prior to physical interaction. At inference time, the network ranks candidate grasps for the specified rearrangement goal, selects the grasp with the highest predicted utility, and executes the unified manipulation policy from the selected configuration. We evaluate the proposed framework on goal-conditioned rearrangement of multiple large household objects in simulation to validate the pipeline's effectiveness in selecting the high-utility grasp configuration for large-object rearrangement.

In summary, our contributions are:
\begin{itemize}
    \item We present a unified goal- and grasp-conditioned loco-manipulation policy trained via PPO that operates across multiple initial grasp configurations for large-object rearrangement.
    
    \item We propose a critic-distillation method where the critic is distilled into a lightweight grasp utility network that predicts downstream grasp utility from pre-interaction grasp and goal information, enabling task-aware grasp selection at inference time.
    
    \item We show that different feasible grasps can lead to substantially different downstream manipulation outcomes, and that our proposed grasp selection framework, Grasp-TAGS, improves task success over fixed or random grasp selection, through simulation experiments
\end{itemize}

\label{sec:01_introduction}

\section{Related Work}

\subsection{Large Object Rearrangement in Loco-Manipulation}
Large-object rearrangement remains a challenging problem in Loco-manipulation. In this work, we refer to large objects as bulky objects that extend beyond the conventional tabletop manipulation setting, such as chairs and carts \cite{SuperQGRASP}. Their size and weight require robots to coordinate base locomotion and end-effector motion to push or pull while firmly holding the object toward a desired pose \cite{fu2022deepwbc, mendonca2024realworldrl, li2025robotmoverlearninglargeobjects}.

Early approaches primarily relied on model-based methods, such as model-predictive control (MPC), to explicitly account for object dynamics and execute whole-body motions analytically in accordance with dynamical laws in physics \cite{Li2022MulticontactMPC, dai2024interactivenavigationadaptivenonprehensile, NavigationAmongMovableObjects, Wang2024AnInteractiveNavigationMethod, MultiContactHeavyObjectPushing}. More recently, data-driven approaches based on reinforcement learning (RL) and imitation learning (IL) have emerged as an alternative to directly handling the complex interaction dynamics of large-object manipulation by learning the knowledge of interaction\cite{alore2026, li2025robotmoverlearninglargeobjects, ravan2024diffusionmobility, mendonca2024realworldrl, Omniretarget_Holosoma, zheng2025embracingbulkyobjectshumanoid, zhang2026sumo, ReLIC}. For instance, Ravan et al. \cite{ravan2024diffusionmobility} develop a teleoperation-based pipeline to collect demonstrations of a quadrupedal robot and subsequently train a diffusion policy \cite{chi2023diffusionpolicy}. In contrast, ALORE \cite{alore2026}, RobotMover \cite{li2025robotmoverlearninglargeobjects}, and ReLIC \cite{ReLIC} adopt hierarchical reinforcement learning to learn manipulation policies in simulation and transfer them to real-world platforms. Such approaches reduce the reliance on costly expert demonstrations and enable scalable policy learning through simulation.

Despite these advances, existing learning-based approaches typically assume a single predefined initial grasp pose (with positional variance smaller than 10 cm) \cite{alore2026, li2025robotmoverlearninglargeobjects, mendonca2024realworldrl, ravan2024diffusionmobility}. This assumption overlooks that larger-scale grasp configurations (positional variance larger than 30 cm) can induce substantially different interaction dynamics, leading to different downstream manipulation performance. Some prior works relax the assumption of a fixed interaction configuration in non-prehensile manipulation. Polverini et al. \cite{object-pushing} exploit multiple robot--environment contacts to push a heavy object, while Dadiotis et al. \cite{MultiContactHeavyObjectPushing} learn to adapt the pushing location and direction when transporting unknown objects toward target poses. However, these works focus on non-prehensile pushing, where contact locations can be adjusted during execution. In contrast, we consider prehensile large-object rearrangement, where the robot must select among spatially distinct candidate grasps on different object parts before manipulation begins and maintain the selected grasp throughout the task.


\subsection{Grasp Planning and Task-Oriented Grasp Selection}

Grasp planning aims to identify end-effector poses that establish stable contact with a target object \cite{SuperQGRASP}. Given that multiple feasible grasps may exist, grasp pose evaluation and selection determine which candidate to execute. Prior approaches mainly evaluate grasps using analytical criteria such as contact forces, object stability, and wrench-based measures \cite{graspit, kumra2020antipodal, grasp-quality-measures, SuperQGRASP}. More recent learning-based methods estimate grasp quality directly from visual observations or point clouds \cite{fang2023anygrasp, sundermeyer2021contactgraspnet, mahler2017dexnet3.0} using expert datasets.

These approaches primarily assess grasp feasibility and stability, without considering their impact on downstream manipulation. Recent task-oriented grasping methods address this limitation by selecting grasps according to the intended manipulation objective \cite{GraspAndManipulation, Grasp-CLIP, WANG202615TGDF, kaynar2023remotetaskorientedgrasparea, DexterousGraspingCritic}. However, these methods predominantly focus on tabletop manipulation, where tasks are relatively short-horizon, and objects are typically smaller and lighter. In contrast, large-object rearrangement requires long-horizon loco-manipulation, where different grasp locations can induce substantially different robot-object configurations and consequently affect downstream task execution \cite{SuperQGRASP, ravan2024diffusionmobility}. In this work, we study task-aware grasp selection for goal-conditioned large-object rearrangement by ranking feasible candidate grasps according to their expected downstream task utility rather than grasp feasibility alone.

\subsection{Conditioned Policy Learning}

Conditioned policy learning incorporates task-level or contextual variables into policy learning, enabling a single policy to adapt its behavior to different task objectives and environments. In imitation learning, policies have been conditioned on variables such as target object poses \cite{chi2023diffusionpolicy, pi0.5-black25a, black2026pi0visionlanguageactionflowmodel}, language instructions \cite{Brohan2022RT1RT, Brohan2023RT2VM, octo_2023, kim24openvla}, and task parameters \cite{Zhao2023ACT}. Reinforcement learning similarly enables policies to condition on task parameters \cite{alore2026, li2025robotmoverlearninglargeobjects, ContextBasedMetaRL}, object physical attributes \cite{wang2026phys2real}, and language instructions \cite{LanguageRL, languageRL2, hu2024flare, nakamoto2024steering}, allowing agents to acquire diverse behaviors through self-exploration rather than expert demonstrations \cite{uvfa, HindsightExperienceReplay, MultiGoalRL, distral, schulman2017ppo}.

Despite the success of conditioned policy learning, initial grasp pose has received limited attention as a conditioning variable for large-object loco-manipulation. A few recent works incorporate grasp pose into imitation-learning-based policies \cite{xiang2026graspldp, PushGraspLearning, foong2026handfulsequentialgraspconditioneddexterous, fang2018tog}, but rely on expert demonstrations to learn the corresponding manipulation behaviors. The work most closely related to ours \cite{DexterousGraspingCritic} incorporates grasp information into value-based grasp evaluation, but focuses on small, geometrically simple objects in tabletop settings. In contrast, we explicitly condition an RL-based loco-manipulation policy on the initial grasp pose, enabling the policy and its value function (i.e., the critic) to capture the distinct interaction dynamics induced by different grasp configurations. We then distill the learned grasp-dependent value estimates into a grasp-selection network that predicts the expected downstream task utility of candidate grasps before executing the policy.

\label{sec:02_related_work}

\section{Problem Statement}
\label{problem-statement}
We consider goal-conditioned large-object rearrangement, where a mobile manipulator moves a target object from an initial pose $p_{0}$ to a desired planar pose $p_{goal}=(x,y,\theta)\in SE(2)$. We assume a set of feasible initial grasp poses $g_1, g_2, \cdots, g_K$, each specifying a desired end-effector pose expressed in the target object's local coordinate frame:
\begin{equation}
    \mathcal{G}=\{g_1,g_2,\ldots,g_K\}\subset SE(3)
\end{equation}

Our objective is twofold. First, we learn a manipulation policy $\pi(a_t|o_t,g_i)$ capable of executing the rearrangement task under different $g_i\in\mathcal{G}$. Second, we select the grasp providing the highest expected downstream task utility:
\begin{equation}
    g^* =
    \arg\max_{g_i\in\mathcal{G}}
    U(g_i,p^{obj}_{goal}),
\end{equation}
where $U(\cdot)$ represents the expected downstream utility of the candidate grasp $g_i$ with respect to the task objective $p^{obj}_{goal}$, which is the target pose represented in the initial object frame. The selected grasp $g^*$ is used for downstream manipulation at inference time:
\begin{equation}
    a_t\sim \pi(\cdot|o_t, g^*)
\end{equation}
\label{sec:03_background_problem_statement}

\section{Methodology}
\begin{figure*}[t]
     \centering
     \includegraphics[ width=2\columnwidth]{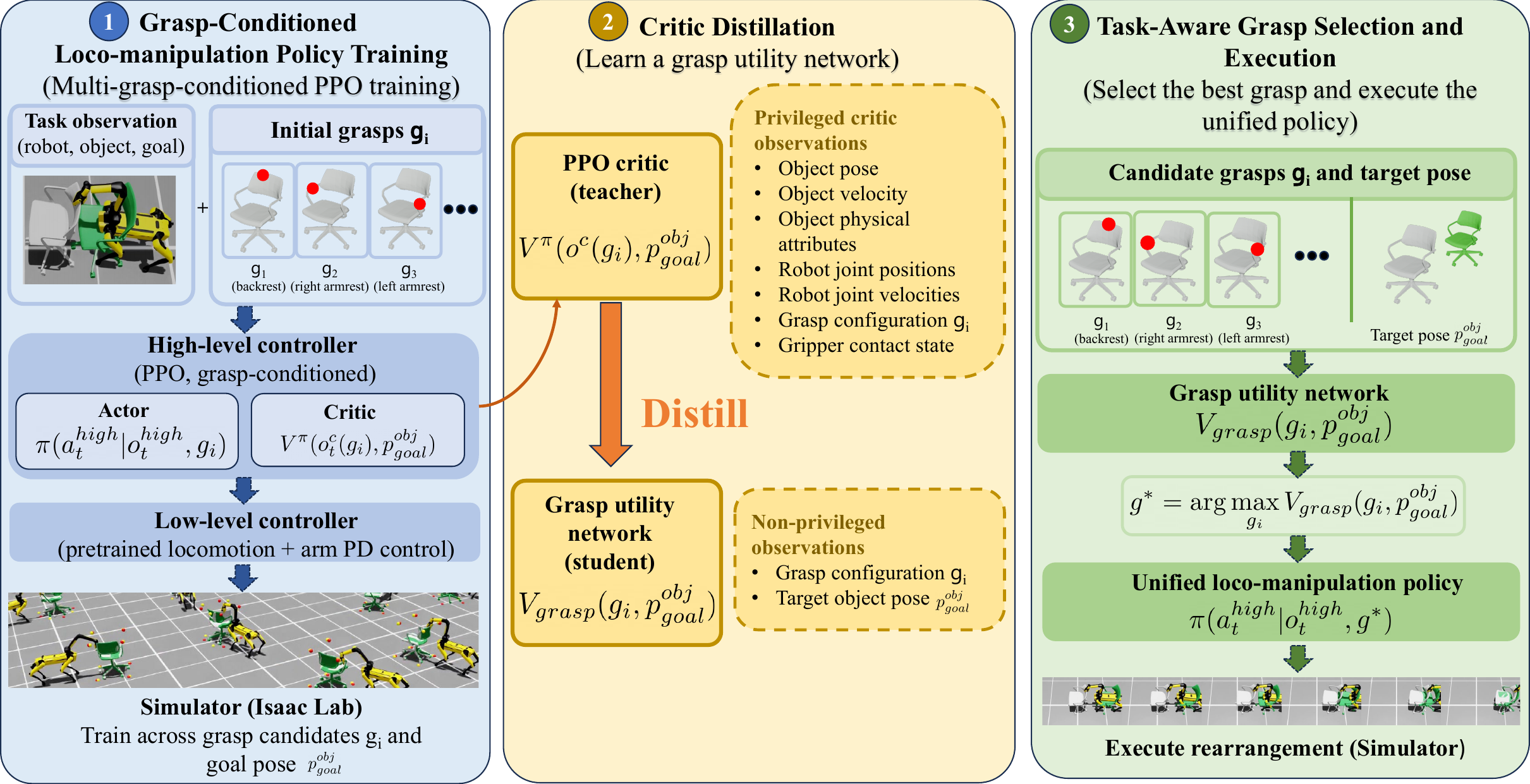}
     \caption{\footnotesize{\textbf{Overview of the proposed pipeline Grasp-TAGS.} (1) A hierarchical policy trained via PPO performs loco-manipulation using a pretrained low-level controller. (2) After PPO training, the learned critic is distilled into a grasp utility network. (3) At inference time, the distilled network selects the grasp $g^*$ with the highest predicted downstream utility before policy execution.}}
     \label{method:hierarchy}
\end{figure*}

Unlike conventional mobile manipulation settings that assume a fixed grasp configuration, we consider multiple feasible grasp poses on the target object. Each grasp pose induces a distinct robot–object interaction configuration and, consequently, different manipulation dynamics. Our objective is two-fold: (1) learn a unified manipulation policy capable of transporting the object to a desired pose under different grasp configurations, and (2) select, at inference time, the grasp pose with the highest expected utility for successfully completing the rearrangement task (see Fig.~\ref{method:hierarchy}).

\subsection{Hierarchical Policy Learning}

Large-object rearrangement requires coordinated locomotion and manipulation over long horizons \cite{fu2022deepwbc}. Directly learning a unified whole-body policy over all robot joints is challenging due to the high-dimensional action space and complex robot-object interactions \cite{alore2026, li2025robotmoverlearninglargeobjects}. To improve learning efficiency, inspired by \cite{ReLIC, kumar2021rma}, we adopt a hierarchical control architecture consisting of a pretrained low-level locomotion controller and a high-level task policy for manipulation. The low-level controller maintains locomotion stability, while the high-level policy generates task-related commands for object rearrangement. This decomposition allows the reinforcement learning agent to focus on task-level manipulation and planning without explicitly learning low-level locomotion dynamics (see Fig.~\ref{method:hierarchy})

\subsubsection{Low-level controller} 
The low-level controller generates dynamically stable locomotion behaviors while tracking task-specific commands from the high-level controller. Specifically, the high-level command is defined as
\begin{equation}
\label{high-level-command}
a^{high}_{t} = (v^b_{x}, v^b_{y}, \omega^b_{z}, \Delta q_{arm, 1:6})
\end{equation}
where $v^b_{x}, v^b_{y}$ are the body linear velocities in the robot base frame, $\omega^b_{z}$ is the body yaw rate in the robot base frame, and $\Delta q_{arm, 1:6}$ are the arm joint angle displacements. Inspired by \cite{ReLIC}, we execute the 6-dim arm commands directly with PD controllers, and we process base velocity commands with a pretrained locomotion policy into leg control commands to maintain stability. The locomotion policy receives the observation
\begin{equation}
o^{low}_{t} = (s^{b}_t, \dot{s}^{b}_t, \mathrm{grav_t}, q_{1:18}, \dot{q}_{1:18},  a^{high}_t, a^{low}_{t-1})
\end{equation}

where $s^{b}_t\in \mathbb{R}^3$
 denotes the observed body pose in robot base frame, $\dot{s}^{b}_t$ denotes the observed body velocity in robot base frame, $\mathrm{grav}_t\in \mathbb{R}^3$
 denotes the projected gravity vector, and $q_{1:18} = (q_{leg, 1:12}, q_{arm, 1:6}), \dot{q}_{1:18} = (\dot{q}_{leg, 1:12}, \dot{q}_{arm, 1:6})\in \mathbb{R}^{18}$
 denote the joint positions and velocities of all robot joints, including the leg joints and the arm joints. In addition, $a^{high}_{t}\in \mathbb{R}^9$
 denotes the high-level command defined in Eq. \ref{high-level-command}. The locomotion policy produces 12-dimensional leg joint commands $a^{low}_t = q_{leg, 1:12}\in \mathbb{R}^{12}$ 
which are executed by the robot to maintain locomotion stability and track the desired base velocity.

\begin{table*}[htbp]
\caption{Reward terms to train the high-level controller}
\label{tab:rewards}
\begin{tabular}{|c|c|c|l|}
\hline
\textbf{Reward Terms}                       & \textbf{Formula}                                                                                            & \textbf{Weight} & \textbf{Description}                                                                                             \\ \hline
Task completion                    & $\mathbb{1}[\|p_o - p_o^*\} < \varepsilon_p$ & 10.0   & Positive reward when the object reaches the target pose                  \\ \hline
Keypoint pose matching             & $\exp\left(-\frac{\| \mathbf{K}_{goal} - \mathbf{K}_{obj} \|_2}{\sigma^2}\right)$                  & 8.0   & Reduce the distance between the current and target object keypoints                \\ \hline
Goal-directed velocity             & $\exp\left(\frac{\hat{\mathbf{v}}_{o} \cdot \hat{\mathbf{d}}_{goal}}{\sigma^2} - 1\right)$    & 1.0   & \makecell[ll]{Encourage alignment between the object velocity and the direction \\toward the goal} \\ \hline
Alive                              & $\mathbb{1}[episode\ is\ alive]$                                                                   & 5.0    & Sparse reward for maintaining a valid and active episode                                                \\ \hline
Linear velocity smoothness         & $\|v_o\|^2$                                                                                        & -2.0   & Penalty on changes in the object's linear velocity                                                      \\ \hline
Angular velocity smoothness        & $\|\omega_o\|^2$                                                                                   & -2.0   & Penalty on changes in the object's angular velocity                                                     \\ \hline
Orientation stability              & $\mathbb{1}[\|g_{o,xy}\|^2>\varepsilon_{orient}]$                                                                                  & -10.0  & Encourage the object to maintain a desired upright/flat orientation                            \\ \hline
Joint configuration regularization & $\|q_{1:6} - q_{default, 1:6}\|$                                                                 & -5.0   & \makecell[ll]{Penalty based on the deviation of the robot joint positions from their \\ default positions      }          \\ \hline
Undesired contact penalty          & $\mathbb{1}[\|F_{o, r}\|>\varepsilon_{contact}]$                                                   & -7.0   & Penalty for undesired contacts during manipulation                                                      \\ \hline
\end{tabular}
\end{table*}

\subsubsection{High-level controller}
\label{high-level-controller}

\begin{table}[htbp]
\caption{Observation terms to the high-level controller. It consists of proprioceptive data from the robot and the real-time observations of the object. }
\label{tab:high-level-obs}
\begin{tabular}{|c|c|c|}
\hline
\textbf{Obs. Terms}          & \textbf{Dim} & \textbf{Description}                                   \\ \hline
$q_{1:128}$          & 18  & Position of all the joints (both arm \& legs) \\ \hline
$\dot{q}_{1:128}$    & 18  & Velocity of all the joints                    \\ \hline
$s^{b}_t$       & 2   & Orientation of the base (no yaw)         \\ \hline
$\dot{s}^{b}_t$ & 3   & Angular velocity of the base                \\ \hline
$a^{high}_{t-1}$    & 9   & Last high-level action                                   \\ \hline
$p^{b}_{ee,t}$      & 7   & SE(3) pose of the end-effector in base frame \\ \hline
$p^{b}_{obj,t}$     & 3   & SE(2) pose of the object in robot base frame       \\ \hline
$p^{obj}_{goal, t}$     & 3   & Target SE(2) pose in object frame             \\ \hline
\end{tabular}
\end{table}

The high-level controller executes a policy focused on task-level planning and manipulation. Inspired by the observation terms in \cite{alore2026}, Table \ref{tab:high-level-obs} summarizes the observations $o^{high}_t$ provided to the high-level controller. The high-level controller outputs the high-level command defined in Eq. \ref{high-level-command}, which the low-level controller then executes.

We train the high-level controller using Proximal Policy Optimization (PPO) \cite{schulman2017ppo}, implemented in RSL-RLproposed \cite{schwarke2025rslrl}, while keeping the pretrained low-level controller fixed throughout training. The objective is to maximize the expected discounted cumulative reward:

\begin{equation}
J(\pi^\theta_{high},\pi_{low}) = \mathbb{E}_{r\sim p(\cdot|\pi^\theta_{high}, \pi_{low})}\left[ \sum_{t=0}^{T-1}\gamma^{t} r_t\right]
\end{equation}

where the reward $r_t$ is composed of the terms described in Sec. \ref{subsec:rewards}. More training details can be found in Sec. \ref{subsec:grasp-conditioned}

\subsection{Grasp-Conditioned Loco-manipulation Policy Training}
\label{subsec:grasp-conditioned}

Rather than learning an independent high-level policy for each grasp, we train a grasp-conditioned policy via PPO \cite{schulman2017ppo} that operates across multiple initial grasp configurations, eliminating the need for separate training for each grasp on the same object \cite{DexterousGraspingCritic}. Following \cite{alore2026}, the arm joint angles implicitly encode the initial grasp pose.

During training, each episode samples one of several predefined feasible grasp poses on the target object. After establishing the grasp, the robot is required to push or pull the object to a desired target pose (x,y,$\theta$), where $x,y\in [-1,1]$ and $\theta \in [-\frac{\pi}{2}, \frac{\pi}{2}]$. The policy therefore learns manipulation behaviors across different initial grasp configurations $g_i$ and rearrangement objectives $p_{goal}$. We implement the training environment in IsaacLab \cite{mittal2025isaaclab} and train the high-level controller using PPO with 4096 parallel environments for 8000 iterations, as described in Sec. \ref{high-level-controller}.

Training a unified policy across multiple grasp poses provides two key advantages. First, it enables the policy to capture the diverse interaction dynamics induced by different robot-object contact configurations without requiring separate training procedures for each grasp pose. Second, the resulting value function provides a common measure of the expected long-horizon manipulation performance across candidate grasp poses, which serves as the foundation for subsequent grasp pose selection (see Sec. \ref{sec:results}).

\subsection{Task-Specific Reward Functions}
\label{subsec:rewards}
We design the reward function for the high-level controller according to three objectives: (i) \textit{task completion}, which encourages the robot to transport the object to the desired pose; (ii) \textit{motion smoothness}, which promotes smooth and non-aggressive manipulation behaviors; and (iii) \textit{interaction stability}, which penalizes undesired robot-object contacts and object toppling. Table \ref{tab:rewards} summarizes the complete set of reward terms.

\subsection{Critic Distillation for Task-Aware Grasp Selection}
\label{subsec:critic-distillation}

As illustrated in Fig.~\ref{method:hierarchy}, the PPO critic \cite{schulman2017ppo} estimates the expected cumulative return from the current critic observation. Thus, the critic implicitly captures the expected downstream utility of different grasp configurations: grasps that enable more successful, efficient, and stable rearrangement should receive higher value estimates. At the beginning of an episode, different initial grasp poses $g_1,g_2,\ldots, g_K$ induce different critic observations $o^c(g_1),o^c(g_2),\ldots, o^c(g_K)$ at the start time $t=0$. Together with the goal pose in the object frame $p^{obj}_{goal}$, the input to the critic can be represented as: 
\begin{equation}
y_i = (o^c(g_i), p^{obj}_{goal})
\end{equation}
Consequently, the critic can provide different value estimates for the downstream task 
\begin{equation}
\label{critic-input}
V^\pi(y_i)
=
\mathbb{E}_{\pi}
\left[
\sum_{t=0}^{T-1}\gamma^t r_t
\mid o^c(g_i), p^{obj}_{goal}
\right].
\end{equation}

However, the critic relies on privileged information unavailable at inference time, including privileged physical and state information, such as object mass and velocity. Although these properties can potentially be estimated through physical interaction \cite{wang2026phys2real, alore2026}, such interaction may require committing to an initial grasp. If that grasp is suboptimal, the robot must release and re-grasp the object, introducing additional manipulation steps and potential failure modes. We therefore avoid explicit physics estimation and instead distill the critic's grasp-dependent value into a lightweight grasp utility network $V_{\mathrm{grasp}}(x_i)$
using only
information on the candidate grasp and the goal object pose:
\begin{equation}
x_i =
\left(g_i, p^{obj}_{goal}\right),
\end{equation}
where $g_i$ represents the candidate
grasp configuration, and $p^{obj}_{goal}$ is the target pose represented in the initial object frame, representing the task objective. By removing privileged physical variables from the input, the grasp selection network learns the expected grasp utility marginalized over the physical-property distribution encountered during training.

After policy training, we sample initial states from all $N$ sub-environments and regress the utility
network toward the corresponding critic values:
\begin{equation}
\mathcal{L}_{\mathrm{distill}}
=
\frac{1}{N}
\sum_{k=1}^{N}
\left[
V_{grasp}(x^{(k)}_{i})
-
V^\pi(y_i^{(k)})
\right]^2.
\end{equation}
where $g^{(k)}_{i}$ is the grasp candidate sampled in the $k$-th sub-environment, $x^{(k)}_{i}$ is the corresponding pre-interaction information and $y^{(k)}_i$ is the associated critic input as defined in Eq. \ref{critic-input}


At inference, for the desired target object pose represented as $p_{goal}^{obj}$, we select
\begin{equation}
g^* =
\arg\max_{g_i\in\mathcal{G}}
V_{grasp}(x_i),
\end{equation}
The manipulation policy is then executed conditioned on the selected grasp pose $g^*$
\label{sec:04_methodology}

\section{Experimental Setup}

We evaluate the proposed grasp-pose-conditioned manipulation policy and critic-distilled grasp pose selection module on large-object rearrangement tasks through quantitative experiments in IsaacLab using synthetic object models.


\subsection{Loco-manipulation Policy Evaluation in Simulation}
\begin{figure}[h]
     \centering
     \includegraphics[width=1\columnwidth]{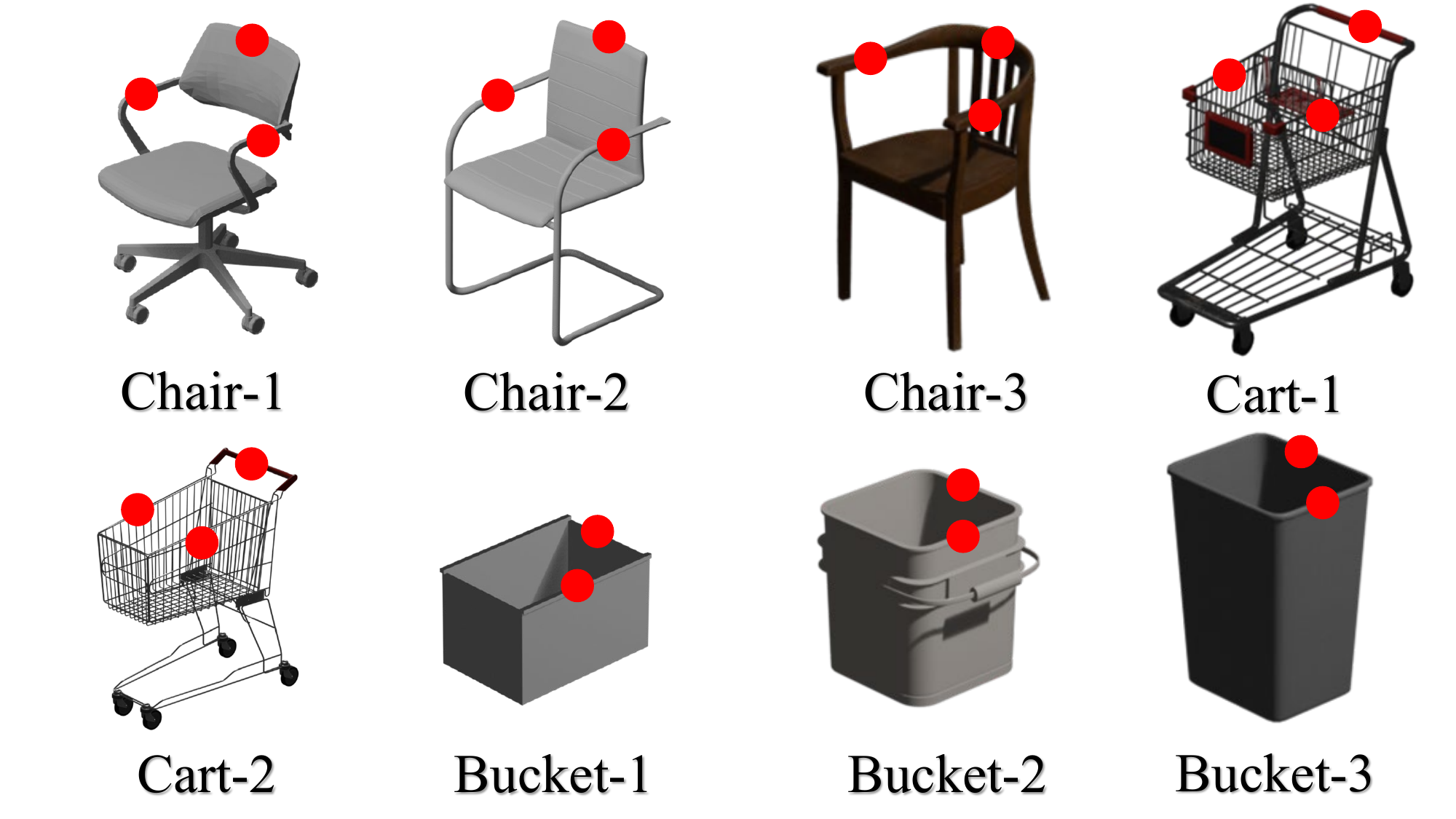}
     \caption{\footnotesize{\textbf{Objects and candidate grasp poses used in the experiments.} The dataset includes three chairs, two carts, and three buckets, with predefined candidate grasps highlighted in red. Candidate grasps used for our experiments are shown in red circle.}}
     \label{data}
\end{figure}

\noindent \textbf{Dataset:}
We construct a dataset of eight common household large objects, including three chairs, two carts, and three buckets, obtained from Fab \cite{fab}. These objects represent common household object categories while covering diverse geometric structures, as shown in Fig.~\ref{data}. For each object, we manually select feasible initial grasp configurations and compute their corresponding inverse kinematics solutions. Specifically:
\begin{itemize}
    \item \textbf{Chairs:} three grasp poses on the two armrests and the backrest.
    \item \textbf{Carts:} three grasp poses on the two sides of the basket and the handle.
    \item \textbf{Buckets:} two grasp poses on the two edges.
\end{itemize}

\noindent \textbf{Large Object Rearrangement Task:} 
As described in Sec.~\ref{problem-statement}, we implement the large-object rearrangement task in IsaacLab \cite{mittal2025isaaclab}. At the beginning of each episode, the target object is placed at the origin and a target $SE(2)$ pose $(x,y,\theta)$ is randomly sampled, where $x,y\in[-1,1]$ and $\theta\in[-\frac{\pi}{2},\frac{\pi}{2}]$. For each object, we train one grasp-conditioned policy across all of its candidate grasps. We evaluate task performance using the metrics described below. We run all experiments in 4096 parallel environments.

\noindent \textbf{Evaluation metrics:} We evaluate the performance of the loco-manipulation policy under a given grasp pose using four metrics after a fixed execution horizon of 750 steps:
\begin{enumerate}[leftmargin=*]
    \item \textbf{Success rate:} the fraction of parallel sub-environments in which the final planar position error is below $0.1$~m, and the final keypoint angular error is below $10^\circ$ \cite{object-pushing};
    \item \textbf{Planar positional error}: measuring the final Euclidean distance on the $xy$-plane between the observed and target object positions;
    \item \textbf{Angular error}: measuring the final absolute yaw difference between the observed and target object orientations.
    \item \textbf{Keypoint Angular Error:} The angular difference between the observed and target object poses estimated from two sets of 2D object keypoints using the Kabsch--Umeyama alignment algorithm \cite{alignment}. We use the eight corners of the object's bounding box as keypoints. This metric provides a more robust estimate of the orientation error than directly comparing the object yaw angles.
\end{enumerate}

\noindent \textbf{Baseline methods:}
We establish three baselines to evaluate the benefits of grasp-pose-conditioned policy learning and the proposed task-aware grasp pose selection module for large-object rearrangement. Specifically:
\begin{enumerate}[leftmargin=*]
    \item \textbf{$PPO_i$:} This baseline consists of policies trained exclusively on each grasp pose $g_i$ (see Fig.\ref{data}) using the same training procedure and hyperparameters. For example, $PPO_1$ is trained from scratch using only grasp pose $g_1$, without exposure to the other candidate grasp poses. These baselines evaluate the effect of training a policy across multiple grasp configurations.

    \item \textbf{$PPO_g+$Random:} This baseline uses the proposed grasp-pose-conditioned loco-manipulation policy trained on all grasp candidates, but does not employ the grasp pose selection module. During both training and inference, a candidate grasp pose is randomly sampled at the beginning of each episode. This baseline isolates the contribution of task-aware grasp pose selection.

    \item \textbf{$PPO_g+$Selection (Grasp-TAGS, Ours):} Our complete pipeline consists of the grasp-pose-conditioned loco-manipulation policy together with the critic-distilled grasp pose selection module. At inference time, the selection module evaluates the candidate grasp poses and selects the one with the highest predicted downstream task utility. This configuration represents the proposed method.
\end{enumerate}


\label{sec:05_experimental_setup}

\section{Results}
\label{sec:results}
\begin{table*}[t]
\centering
\caption{\textbf{Quantitative results on simulated large-object rearrangement.} We compare $PPO_i$, $PPO_g+$Random, and Grasp-TAGS using Success rate, Planar Positional Error, Angular Error and Keypoint Angular Error across multiple objects ($\downarrow$: smaller numbers indicate better results; $\uparrow$: larger numbers indicate better results).}
\label{tab:quantitative}
\resizebox{\textwidth}{!}{
\begin{tabular}{|c|ccc|ccc|}
\hline
                        & \multicolumn{3}{c|}{\bf{Chair-1}}                                                                                 & \multicolumn{3}{c|}{\bf{Chair2}}                                                                                  \\ \hline
                        & \multicolumn{1}{c|}{\bf{$PPO_i$}}                        & \multicolumn{1}{c|}{\bf{$PPO_g$ + Random}} & \textbf{Grasp-TAGS}      & \multicolumn{1}{c|}{\bf{$PPO_i$}}                        & \multicolumn{1}{c|}{\bf{$PPO_g$ + Random}} & \textbf{Grasp-TAGS}       \\ \hline
\textbf{Success rate [$\%$] ($\uparrow$)}            & \multicolumn{1}{c|}{70.09/75.46/78.20}             & \multicolumn{1}{c|}{75.88}           & \textbf{93.65}  & \multicolumn{1}{c|}{63.79/71.51/70.00}              & \multicolumn{1}{c|}{60.28}           & \textbf{84.72}   \\ \hline
\textbf{Planar Positional Error [m] ($\downarrow$)} & \multicolumn{1}{c|}{0.0292/0.0250/0.0366}          & \multicolumn{1}{c|}{0.0281}          & \textbf{0.0162} & \multicolumn{1}{c|}{0.034/\textbf{0.0161}/0.0142}           & \multicolumn{1}{c|}{0.045}           & 0.0239           \\ \hline
\textbf{Angular Error [rad] ($\downarrow$)}           & \multicolumn{1}{c|}{0.3571/0.2850/0.2802}          & \multicolumn{1}{c|}{0.2407}          & \textbf{0.0712} & \multicolumn{1}{c|}{0.5268/0.3421/0.3732}          & \multicolumn{1}{c|}{0.4382}          & \textbf{0.1308}  \\ \hline
\textbf{Keypoint Angular Error [deg] ($\downarrow$)}  & \multicolumn{1}{c|}{20.4597/16.3271/16.0567}       & \multicolumn{1}{c|}{13.7906}         & \textbf{4.0793} & \multicolumn{1}{c|}{30.1858/19.5889/21.3821}       & \multicolumn{1}{c|}{25.1091}         & \textbf{7.4923}  \\ \hline
                        & \multicolumn{3}{c|}{\textbf{Chair-3}}                                                                                 & \multicolumn{3}{c|}{\textbf{Cart-1}}                                                                                   \\ \hline
                        & \multicolumn{1}{c|}{\bf{$PPO_i$}}                        & \multicolumn{1}{c|}{\bf{$PPO_g$ + Random}} & \textbf{Grasp-TAGS}      & \multicolumn{1}{c|}{\bf{$PPO_i$}}                        & \multicolumn{1}{c|}{\bf{$PPO_g$ + Random}} & \textbf{Grasp-TAGS}       \\ \hline
\textbf{Success rate [$\%$] ($\uparrow$)}            & \multicolumn{1}{c|}{75.93/68.41/72.22}             & \multicolumn{1}{c|}{72.49}           & \textbf{89.62}  & \multicolumn{1}{c|}{82.13/68.48/67.38}             & \multicolumn{1}{c|}{84.62}           & \textbf{95.51}   \\ \hline
\textbf{Planar Positional Error [m] ($\downarrow$)} & \multicolumn{1}{c|}{\textbf{0.0204}/0.0626/0.0307} & \multicolumn{1}{c|}{0.041}           & 0.0309 & \multicolumn{1}{c|}{0.0291/0.0989/0.0924} & \multicolumn{1}{c|}{0.0426}          & \textbf{0.0283}  \\ \hline
\textbf{Angular Error [rad] ($\downarrow$)}           & \multicolumn{1}{c|}{0.3546/0.4425/0.3628}          & \multicolumn{1}{c|}{0.3616}          & \textbf{0.1212} & \multicolumn{1}{c|}{0.1716/0.2522/0.3781}          & \multicolumn{1}{c|}{0.1471}          & \textbf{0.0575}  \\ \hline
\textbf{Keypoint Angular Error [deg] ($\downarrow$)}  & \multicolumn{1}{c|}{20.3148/25.3516/20.7884}       & \multicolumn{1}{c|}{20.7177}         & \textbf{6.9462} & \multicolumn{1}{c|}{9.8312/14.5265/20.9177}        & \multicolumn{1}{c|}{8.4009}          & \textbf{3.2925}  \\ \hline
                        & \multicolumn{3}{c|}{\textbf{Cart-2}}                                                                                  & \multicolumn{3}{c|}{\textbf{Bucket-1}}                                                                                 \\ \hline
                        & \multicolumn{1}{c|}{\bf{$PPO_i$}}                        & \multicolumn{1}{c|}{\bf{$PPO_g$ + Random}} & \textbf{Grasp-TAGS}      & \multicolumn{1}{c|}{\bf{$PPO_i$}}                        & \multicolumn{1}{c|}{\bf{$PPO_g$ + Random}} & \textbf{Grasp-TAGS}       \\ \hline
\textbf{Success rate [$\%$] ($\uparrow$)}            & \multicolumn{1}{c|}{62.96/81.32/72.75}             & \multicolumn{1}{c|}{64.38}           & \textbf{86.89}  & \multicolumn{1}{c|}{88.89/86.18}                   & \multicolumn{1}{c|}{81.86}           & \textbf{92.58}   \\ \hline
\textbf{Planar Positional Error [m] ($\downarrow$)} & \multicolumn{1}{c|}{0.0901/\textbf{0.0235}/0.0762} & \multicolumn{1}{c|}{0.1158}          & 0.0551 & \multicolumn{1}{c|}{\textbf{0.0101}/0.0280}        & \multicolumn{1}{c|}{0.0201}          & 0.0138  \\ \hline
\textbf{Angular Error [rad] ($\downarrow$)}           & \multicolumn{1}{c|}{0.4028/0.3335/0.3844}          & \multicolumn{1}{c|}{0.371}           & \textbf{0.1321} & \multicolumn{1}{c|}{0.1428/0.2250}                 & \multicolumn{1}{c|}{0.2455}          & \textbf{0.1279}  \\ \hline
\textbf{Keypoint Angular Error [deg] ($\downarrow$)}  & \multicolumn{1}{c|}{23.0807/19.1110/22.0228}       & \multicolumn{1}{c|}{21.2587}         & \textbf{7.5734} & \multicolumn{1}{c|}{8.1790/12.8935}                & \multicolumn{1}{c|}{14.0673}         & \textbf{7.3268}  \\ \hline
                        & \multicolumn{3}{c|}{\textbf{Bucket-2}}                                                                                & \multicolumn{3}{c|}{\textbf{Bucket-3}}                                                                                 \\ \hline
                        & \multicolumn{1}{c|}{\bf{$PPO_i$}}                        & \multicolumn{1}{c|}{\bf{$PPO_g$ + Random}} & \textbf{Grasp-TAGS}      & \multicolumn{1}{c|}{\bf{$PPO_i$}}                        & \multicolumn{1}{c|}{\bf{$PPO_g$ + Random}} & \textbf{Grasp-TAGS}       \\ \hline
\textbf{Success rate [$\%$] ($\uparrow$)}            & \multicolumn{1}{c|}{76.10/81.08}                   & \multicolumn{1}{c|}{82.23}           & \textbf{91.87}  & \multicolumn{1}{c|}{79.91/70.00}                   & \multicolumn{1}{c|}{70.85}           & \textbf{83.89}   \\ \hline
\textbf{Planar Positional Error [m] ($\downarrow$)} & \multicolumn{1}{c|}{0.0219/0.0220}        & \multicolumn{1}{c|}{0.0144}          & \textbf{0.0105} & \multicolumn{1}{c|}{\textbf{0.0258} /0.0265}       & \multicolumn{1}{c|}{0.0689}          & 0.029   \\ \hline
\textbf{Angular Error [rad] ($\downarrow$)}           & \multicolumn{1}{c|}{0.3370/0.2490}                 & \multicolumn{1}{c|}{0.2822}          & \textbf{0.1343} & \multicolumn{1}{c|}{0.2598/0.3199}                 & \multicolumn{1}{c|}{0.3609}          & \textbf{0.2276}  \\ \hline
\textbf{Keypoint Angular Error [deg] ($\downarrow$)}  & \multicolumn{1}{c|}{19.3076/14.2670}               & \multicolumn{1}{c|}{16.1711}         & \textbf{7.6976} & \multicolumn{1}{c|}{14.8877/18.3309}               & \multicolumn{1}{c|}{20.6764}         & \textbf{13.0426} \\ \hline
\end{tabular}
}
\end{table*}

\begin{figure*}[htbp!]
     \centering
     \includegraphics[width=\textwidth]{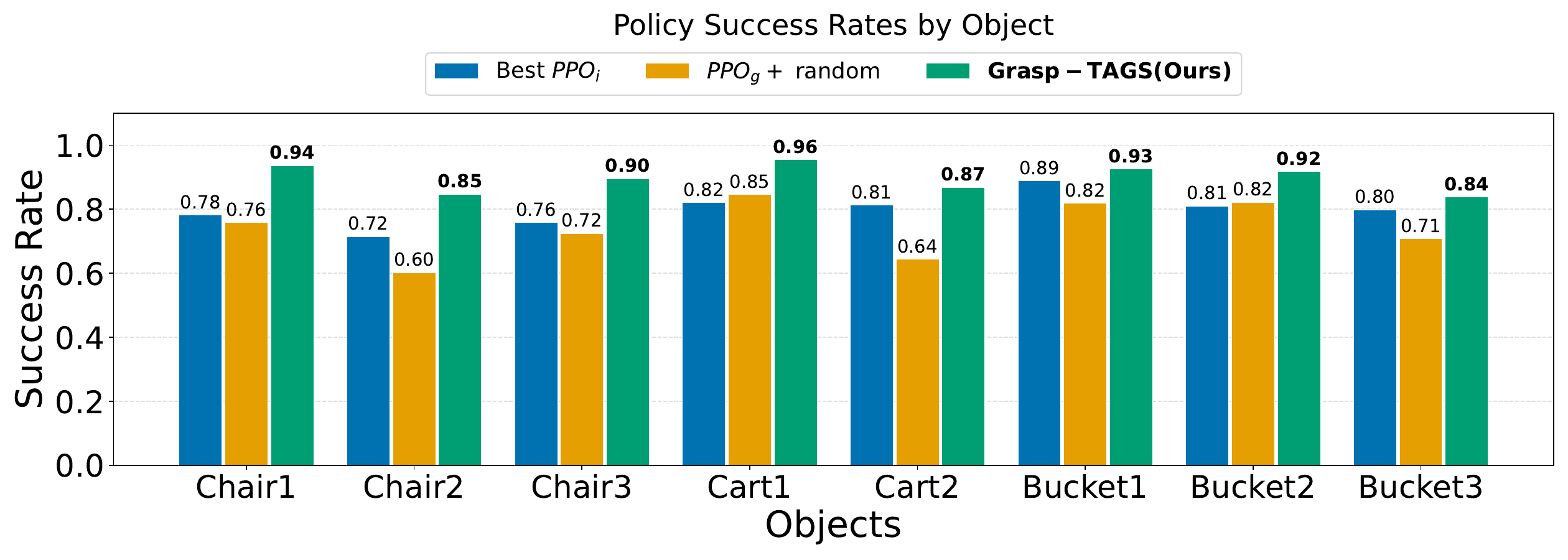}
     \caption{\footnotesize{\textbf{Success rates} for the best $PPO_i$ among the policies trained individually on each grasp pose, $PPO_g+$random grasp selection, and Grasp-TAGS. Grasp-TAGS achieves the highest \textbf{success rate ($\uparrow$)} across all evaluated objects in all 4096 parallel environments.}}
     \label{fig:success_rates}
\end{figure*}
\subsection{Loco-manipulation Policy Evaluation in Simulation}

Our quantitative results indicate that the initial grasp configuration substantially affects downstream loco-manipulation performance (see Table~\ref{tab:quantitative}). Individually trained $PPO_i$ policies can outperform $PPO_g+$Random by achieving a higher success rate and lower positional or Angular Errors, because each policy specializes in a single robot-object interaction configuration. However, the relative performance of different grasps varies across rearrangement goals, motivating explicit task-aware grasp selection.

Combining the unified policy with the proposed grasp selection network improves average task success across all sub-environments (see Fig.~\ref{fig:success_rates}). For Chair-1, Grasp-TAGS achieves a success rate of $93.65\%$ over all sub-environments compared with $75.88\%$ for random grasp selection and $69.02$--$78.20\%$ for the individual single-grasp policies. The proposed method also achieves around a $ 60\%$ lower Angular Error, suggesting that selecting an appropriate interaction configuration is particularly beneficial for tasks that require coordinated object translation and rotation. The same trend is observed for the other objects.

\begin{figure}[h]
     \centering
     \includegraphics[width=0.7\columnwidth]{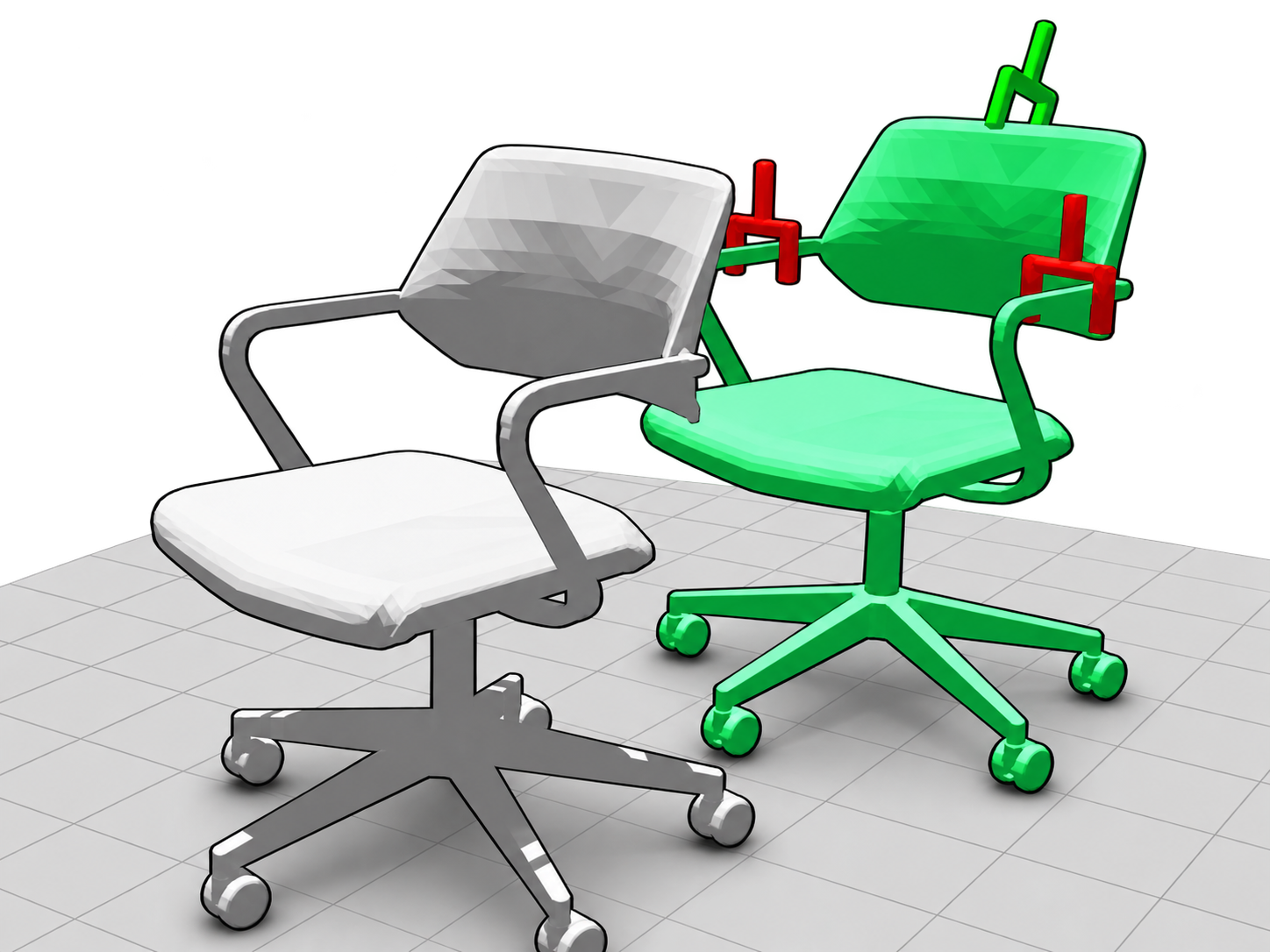}
     \caption{\footnotesize{\textbf{Illustrative example of grasp utility predictions for a target object pose (green mesh)}. The colors visualize the relative utility estimated by Grasp-TAGS, where green denotes higher predicted downstream utility and red denotes lower predicted utility.}}
     \label{qualitative2}
\end{figure}

\noindent \textbf{Quantitative Results:} 
The quantitative results lead to the following observations:
\begin{enumerate}[leftmargin=*]
    \item \textbf{Success rate ($\uparrow$)}: Grasp-TAGS achieves the highest success rate across all eight evaluated objects. Compared with the best-performing single-grasp policy $PPO_i$, Grasp-TAGS improves success by 3.69--15.45 percentage points, and compared with $PPO_g+$Random, by 9.64--24.44 percentage points. Averaged across all objects, Grasp-TAGS achieves a success rate of 89.84$\%$, compared with 79.87$\%$ for the best $PPO_i$ baseline and 74.07$\%$ for $PPO_g+$Random. These results provide quantitative evidence that selecting the initial grasp according to predicted downstream task utility improves the reliability of large-object rearrangement.

    \item \textbf{\textbf{Planar Positional Error ($\downarrow$)}:} Grasp-TAGS achieves the lowest planar positional error on Chair-1, Cart-1, and Bucket-2, and outperforms $PPO_g+$Random on all eight evaluated objects. On the remaining objects, the gap to the best single-grasp policy is small for Chair-2, Chair-3, Bucket-1, and Bucket-3, ranging from approximately 0.32 to 1.05 cm. The largest gap occurs on Cart-2, where Grasp-TAGS is 3.16 cm higher than the best $PPO_i$ baseline. Averaged across all objects, Grasp-TAGS achieves a planar positional error of 2.60 cm, compared with 2.12 cm for the best single-grasp policies and 4.70 cm for $PPO_g+$Random. 
    
    \item {Angular Error and Keypoint Angular Error:} Grasp-TAGS achieves the lowest error under both orientation metrics across all eight evaluated objects. Compared with the strongest baseline for each object, the mean Angular Error decreases from 0.259 rad to 0.125 rad, corresponding to a reduction of approximately 51.6$\%$. Similarly, the mean Keypoint Angular Error decreases from 14.82 to 7.18, a reduction of approximately 51.5$\%$. These results indicate that task-aware grasp selection is particularly effective at improving final object orientation alignment with the desired target pose.

    \item \textbf{Effect of Unified Policy Learning:} The single-grasp policies $PPO_i$ often outperform $PPO_g+$Random because each policy specializes in a single grasp configuration, whereas the unified policy must accommodate a broader range of initial grasp conditions. Nevertheless, when combined with the proposed grasp selection module, the unified policy consistently achieves higher success rates than the best single-grasp baseline across all eight evaluated objects. On average, Grasp-TAGS improves the success rate from 79.87$\%$ for the best $PPO_i$ baseline to 89.84$\%$, corresponding to an absolute improvement of 9.97 percentage points. This result suggests that task-aware grasp selection can compensate for, and ultimately exceed, the performance advantage of individually specialized single-grasp policies.
\end{enumerate}

These results suggest that grasp selection should consider downstream task utility rather than grasp stability alone \cite{graspit}. In loco-manipulation, the initial grasp influences how effectively the robot controls the object throughout execution. Accordingly, our grasp selection network predicts the utility of candidate grasps and selects the most suitable one for the downstream task (Fig.~\ref{qualitative2}), improving overall rearrangement performance. For more qualitative results, please refer to our webpage: \url{https://rpm-lab-umn.github.io/Grasp-TAGS-web-page/}
\label{sec:06_results}

\section{Conclusion}
We presented a task-aware grasp selection framework for prehensile goal-conditioned large-object loco-manipulation. By training one unified manipulation policy across multiple initial grasp configurations, the learned critic captures the long-horizon utility of different grasp configurations. Distilling these values into a lightweight grasp utility network enables the robot to select a suitable grasp before physical interaction without explicitly estimating privileged physical properties of the object. Simulation results demonstrate improved rearrangement success over fixed and randomly selected grasp configurations. Future work will extend the evaluation to additional household objects and real-world deployment on the SPOT platform.
\label{sec:07_conclusion}

\section*{Acknowledgement}
The authors would like to thank Ebasa Temesgen for helps with the initial setup of Isaac Lab environment and all other members of the Robotics: Perception and Manipulation (RPM) Lab for their insightful discussions and proofreading of this paper. 

\bibliographystyle{IEEEtran}
\bibliography{refs}

\end{document}